\pdfoutput=1
\documentclass[11pt]{article}
\usepackage{scalefnt}
\usepackage[]{EMNLP2022}
\usepackage{times}
\usepackage{latexsym}
\usepackage{graphicx}
\usepackage[T1]{fontenc}
\usepackage[utf8]{inputenc}
\usepackage{microtype}
\usepackage{inconsolata}

\usepackage{todonotes}

\title{Extended Multilingual Protest News Detection - \\ Shared Task 1, CASE 2021 and 2022}

\author{Ali Hürriyetoğlu  \\
  {\small KNAW Humanities Cluster DHLab} \\
  \texttt{\small ali.hurriyetoglu@dh.huc.knaw.nl} \\\And
  Osman Mutlu \\
  {\small Koc University} \\
  \texttt{\small omutlu@ku.edu.tr} \\\And
  Fırat Duruşan \\
  {\small Koc University} \\
  \texttt{\small fdurusan@ku.edu.tr} \\\AND
  Onur Uca \\
  {\small Mersin University} \\
  \texttt{\small onuruca@mersin.edu.tr} \\\And
  Alaeddin Selçuk G{\"u}rel \\
  {\small Huawei} \\
  \texttt{\small alaeddinselcukgurel@gmail.com} \\\And
  Benjamin Radford \\
  {\small UNC Charlotte} \\
  \texttt{\small benjamin.radford@uncc.edu} \\\AND
  Yaoyao Dai \\
  {\small UNC Charlotte} \\
  \texttt{\small yaoyao.dai@uncc.edu} \\\And
  Hansi Hettiarachchi \\
  {\small Birmingham City University} \\
  \texttt{\small hansi.hettiarachchi@mail.bcu.ac.uk} \\\And
  Niklas Stoehr \\
  {\small ETH Zurich} \\
  \texttt{\small niklas.stoehr@inf.ethz.ch} \\\AND
  Tadashi Nomoto \\
  {\small National Institute of Japanese Literature} \\
  \texttt{\small nomoto@acm.org} \\\And
  Milena Slavcheva \\
  {\small Bulgarian Academy of Sciences} \\
  \texttt{\small milena@lml.bas.bg} \\\And
  Francielle Vargas  \\
  {\small University of São Paulo} \\
  \texttt{\small francielleavargas@usp.br} \\\AND
  Aaqib Javid \\
  {\small Koc University} \\
  \texttt{\small ajavid20@ku.edu.tr} \\\And
  Fatih Beyhan \\
  {\small Sabanci University} \\
  \texttt{\small fatihbeyhan@sabanciuniv.edu} \\\And
  Erdem Yörük \\
  {\small Koc University} \\
  \texttt{\small eryoruk@ku.edu.tr}}

\begin{document}
\maketitle
\vspace*{135px}
\begin{abstract}
We report results of the CASE 2022 Shared Task 1 on Multilingual Protest Event Detection. This task is a continuation of CASE 2021 that consists of four subtasks that are i) document classification, ii) sentence classification, iii) event sentence coreference identification, and iv) event extraction. The CASE 2022 extension consists of expanding the test data with more data in previously available languages, namely, English, Hindi, Portuguese, and Spanish, and adding new test data in Mandarin, Turkish, and Urdu for Sub-task 1, document classification. The training data from CASE 2021 in English, Portuguese and Spanish were utilized. Therefore, predicting document labels in Hindi, Mandarin, Turkish, and Urdu occurs in a zero-shot setting. The CASE 2022 workshop accepts reports on systems developed for predicting test data of CASE 2021 as well. We observe that the best systems submitted by CASE 2022 participants achieve between 79.71 and 84.06 F1-macro for new languages in a zero-shot setting. The winning approaches are mainly ensembling models and merging data in multiple languages. The best two submissions on CASE 2021 data outperform submissions from last year for Subtask 1 and Subtask 2 in all languages. Only the following scenarios were not outperformed by new submissions on CASE 2021: Subtask 3 Portuguese \& Subtask 4 English.
\end{abstract}
\vspace*{150px}

\section{Introduction}
We aim at determining event trigger and its arguments in a text snippet in the scope of an event extraction task. The performance of an automated system depends on the target event type as it may be broad or potentially the event trigger(s) can be ambiguous. The context of the trigger occurrence may need to be handled as well. For instance, the `protest' event type may be synonymous with `demonstration' or not in a specific context. Moreover, the hypothetical cases such as future protest plans may need to be excluded from the results. Finally, the relevance of a protest depends on the actors as in a contentious political event only citizen-led events are in the scope. This challenge is even harder in a cross-lingual and zero-shot setting in case training data are not available in new languages. 

We provide a benchmark that consists of four subtasks and multiple languages in the scope of the 5th Workshop on Challenges and Applications of Automated Extraction of Socio-political Events from Text at The 2022 Conference on Empirical Methods in Natural Language Processing (CASE @ EMNLP 2022)~\cite{hurriyetoglu-etal-2022-challenges}.\footnote{\url{https://emw.ku.edu.tr/case-2022/}, accessed on November 13, 2022.} bgenfrhumil: To paraphrase: The work presented in this paper is a continuation of the work initiated in CASE 2021 Task 1~\cite{hurriyetoglu-etal-2021-multilingual} and consists in adding new documents in already available languages, as well as adding new languages to the evaluation data.

Task 1 consists of the following subtasks that ensure the task is tackled incrementally: i) Document classification, ii) Sentence classification, iii) event sentence coreference identification, and iv) event extraction. The training data consist of documents in English, Portuguese, and Spanish, while the evaluation texts are in English, Hindi, Mandarin, Portuguese, Spanish, Turkish, and Urdu. Subtask 1 ensures documents with relevant senses of event triggers are selected. Next, Subtask 2 focuses on identifying event sentences in a document. Discriminating sentences that are about separate events and grouping them is done in Subtask 3~\cite{hurriyetoglu-etal-2020-automated,hurriyetoglu-et-al-2022-event}. Finally, the sentences that are about the same events are processed to identify the event trigger and its arguments in Subtask 4. In addition to accomplishing the event extraction task, the subtask division improves significantly the annotation quality, as the annotation team can focus on a specific part of the task and errors in previous levels are corrected during the preparation of the following subtask~\cite{hurriyetoglu-etal-2021-cross}. The significance of this specific task division is twofold: i) facilitating the work with a random sample of documents by first identifying relevant documents and sentences before annotating or processing a sample or a complete archive of documents respectively; ii) increasing the generalizability of the automated systems that may be developed using this data~\cite{yoruk-etal-2021-random,mutlu-2022-utilizing}. 

The current report is about Task 1 in the scope of CASE 2022. Task 2~\cite{zavarella-etal-2022-covid19} and Task 3~\cite{tan-EtAl:2022:LREC,tan-etal-2022-event} complement Task 1 by evaluating Task 1 systems on events related to COVID-19 and detecting causality respectively. 

The following section, which is Section~\ref{sec:data} describes the data we use for the shared task. Next we describe the evaluation setting in Section~\ref{sec:evaluation}. The results are provided in Section~\ref{sec:results}. Finally, the Section~\ref{sec:conclusion} conclude this report.

\section{Data}
\label{sec:data}

We used the CASE 2021 training data as those for CASE 2022.\footnote{\url{https://github.com/emerging-welfare/case-2021-shared-task} for CASE 2021 and \url{https://github.com/emerging-welfare/case-2022-multilingual-event} for CASE 2022.} The CASE 2022 test data are the union of CASE 2021 test data and additional new documents in both available and new languages. The new languages are Mandarin, Turkish, and Urdu. 

The new document level data, which are used to extend CASE 2021 data, were randomly sampled from MOT v1.2 ~\cite{palen-michel-etal-2022-multilingual} \footnote{\url{https://github.com/bltlab/mot}} and were annotated by co-authors of this report. Documents were annotated by native speakers of the respective language. A single label was attached to each document. The annotation manual followed in the annotation process~\cite{durusan-et-al-2022-global} was the same as that used in CASE 2021.

The total number of CASE 2022 documents with labels is 3,870 for English, 267 for Hindi, 300 for Mandarin, 670 for Portuguese, 399 for Spanish, 300 for Turkish, and 299 for Urdu.

Teams that developed systems for Subtasks 2, 3, and 4 evaluated their systems on CASE 2021 test data. 

\section{Evaluation setting}
\label{sec:evaluation}

We utilized Codalab for evaluation of Task 1 for CASE 2022.\footnote{\url{https://codalab.lisn.upsaclay.fr/competitions/7438}, accessed on November 13, 2022.} The evaluation for CASE 2021 was performed on an additional scoring page\footnote{\url{https://codalab.lisn.upsaclay.fr/competitions/7126}, accessed on November 13, 2022.} of the original\footnote{\url{https://competitions.codalab.org/competitions/31247}, which is not accessible due to change of the servers of Codalab.} CASE 2021 Codalab page. Moreover, we launched an additional scoring page for CASE 2022 after completion of the official evaluation period.\footnote{\url{https://codalab.lisn.upsaclay.fr/competitions/7768}, accessed on November 13, 2022.}

Five submissions per subtask and language pair could be submitted in total for CASE 2022. The additional scoring phase of both CASE 2021 and CASE 2022 allow only one submission per subtask and language combination per day. The test data of CASE 2021 were shared with participants at the same time with the training data. But the CASE 2022 evaluation data were shared around two weeks before the deadline for submission.

The same evaluation scores that are F1-macro for Subtasks 1 and 2, CoNLL-2012\footnote{\url{https://github.com/LoicGrobol/scorch}, accessed on November 13, 2022.} for Subtask 3, and CoNLL-2000\footnote{\url{https://github.com/sighsmile/conlleval}, accessed on November 13, 2022.} script for Subtask 4 were utilized.

\section{Results}
\label{sec:results}

Eighteen teams were registered for the task and obtained the training and test data for both CASE 2022 and CASE 2021. Ten and seven teams submitted their results for CASE 2021 and CASE 2022 respectively. Seven papers were submitted as system description papers to the CASE 2022 workshop in total. The scores of the submissions are calculated on two different Codalab pages for CASE 2021\footnote{\url{https://codalab.lisn.upsaclay.fr/competitions/7126\#results}, accessed on Nov 14, 2022.} and CASE 2022\footnote{\url{https://codalab.lisn.upsaclay.fr/competitions/7438\#results}, accessed on Nov 14, 2022.}. The teams that have participated are ARC-NLP~\cite{sahin-etal-2022-arc-nlp}, CamPros~\cite{kumari-etal-2022-campros}, CEIA-NLP~\cite{fernandes-etal-2022-ceia}, ClassBases~\cite{wiriyathammabhum-2022-classbases}, EventGraph~\cite{you-etal-2022-eventgraph}, NSUT-NLP~\cite{suri-etal-2022-nsutnlp}, SPARTA~\cite{muller-and-dafnos-2022-sparta}. We provide details of the results and submissions of the participating teams for each subtask in the following subsections.\footnote{The results and system descriptions from participants that did not submit a system description paper are provided as well. This shows the capacity of the state-of-the-art systems on our benchmark. These systems are provided with their codalab names that are colabhero, fine\_sunny\_day, gauravsingh, lapardnemihk9989, lizhuoqun2021\_iscas.}


\begin{table*}[!th]
\centering
\scalefont{0.75}
\begin{tabular}{llllllll}
\hline
Team &   English &  Portuguese & Spanish & Hindi & Turkish & Urdu & Mandarin  \\
\hline
ARC-NLP & 80.74\textsubscript{4} & 79.85\textsubscript{2} & 69.44\textsubscript{5} & 80.08\textsubscript{4} & 84.06\textsubscript{1} & 77.99\textsubscript{3} & 83.39\textsubscript{1} \\
CEIA-NLP & 80.77\textsubscript{3} & 80.07\textsubscript{1} & 73.19\textsubscript{3} & 78.17\textsubscript{6} & 82.43\textsubscript{4} & 77.65\textsubscript{4} & 77.63\textsubscript{4} \\
CamPros & 76.52\textsubscript{7} & 77.11\textsubscript{6} & 69.55\textsubscript{4} & 80.49\textsubscript{2} & 74.75\textsubscript{6} & 73.77\textsubscript{6} & 75.90\textsubscript{6} \\
ClassBases & 78.50\textsubscript{6} & 77.11\textsubscript{5} & 69.25\textsubscript{6} & 80.78\textsubscript{1} & 78.57\textsubscript{5} & 75.72\textsubscript{5} & 77.16\textsubscript{5} \\
NSUT-NLP & 80.62\textsubscript{5} & 73.02\textsubscript{7} & 64.45\textsubscript{7} & 56.71\textsubscript{7} & 67.02\textsubscript{7} & 65.55\textsubscript{7} & 75.45\textsubscript{7} \\
fine\_sunny\_day & 82.22\textsubscript{2} & 79.05\textsubscript{4} & 73.84\textsubscript{2} & 80.11\textsubscript{3} & 82.91\textsubscript{2} & 79.71\textsubscript{1} & 80.99\textsubscript{3} \\
lizhuoqun2021\_iscas & 82.49\textsubscript{1} & 79.22\textsubscript{3} & 74.96\textsubscript{1} & 80.01\textsubscript{5} & 82.89\textsubscript{3} & 78.67\textsubscript{2} & 83.06\textsubscript{2} \\
\hline
\end{tabular}
\caption{The performance of the submissions in terms of F1-macro and their ranks as a subscript for each language and each team participating in CASE 2022 subtask 1.}
\label{CASE2022subtask1results}
\end{table*}

\subsection{CASE 2022 Subtask 1}
The results for CASE 2022 subtask 1 are provided in Table ~\ref{CASE2022subtask1results}. 
ARC-NLP finetune an ensemble of transformer-based language models and use ensemble learning, varying training data for each target language. They also perform tests with automatic translation of both training and test sets. They achieve 1st place both in Turkish and Mandarin, 2nd place in Portuguese and 3rd to 5th place in other languages.
CEIA-NLP finetune XLM-Roberta-base transformers model with all the training data to achieve 1st place in Portuguese, 3rd or 4th places in other languages.
ClassBases achieve 1st place in Hindi test data finetuning XLM-Roberta-large model, 5th or 6th places in other languages.

CamPros finetune XLM-Roberta-base model with all training data, and NSUT-NLP finetune mBERT while augmenting the data by translating different languages into each other.

\subsection{CASE 2021 Subtask 1}
The extended results for CASE 2021 subtask 1 are provided in Table ~\ref{CASE2021subtask1results}. The boldness indicates CASE 2022 entries. 
ClassBases finetune XLM-Roberta-large transformers model to perform 1st in Hindi and 2nd in Portuguese test data. They also achieve 5th and 6th places in Spanish and English respectively. Another team that submitted their model to CASE 2021 test data is ARC-NLP, taking 5th, 8th and 9th places in Portuguese, Spanish and English.

\begin{table}[!th]
\centering
\scalefont{0.85}
\resizebox{\columnwidth}{!}{\begin{tabular}{lllll}
\hline
Team &   English &  Hindi & Portuguese & Spanish  \\
\hline
ALEM                   &  80.82\textsubscript{10} &  N/A & 72.98\textsubscript{11} & 46.47\textsubscript{13} \\
AMU-EuraNova           &  53.46\textsubscript{15} & 29.66\textsubscript{11} & 46.47\textsubscript{14} & 46.47\textsubscript{13} \\
DAAI                   &  84.55\textsubscript{3} & 77.07\textsubscript{6} & 82.43\textsubscript{4} & 69.31\textsubscript{10} \\
DaDeFrTi               &  80.69\textsubscript{11} & 78.77\textsubscript{3} & 77.22\textsubscript{10} & 73.01\textsubscript{7} \\
FKIE\_itf\_2021        &  73.90\textsubscript{13} & 54.24\textsubscript{10} & 62.39\textsubscript{12} & 68.20\textsubscript{11} \\
HSAIR &  77.58\textsubscript{12} & 59.55\textsubscript{9} & 81.21\textsubscript{7} & 69.84\textsubscript{9} \\
IBM MNLP IE            &  83.93\textsubscript{4} & 78.53\textsubscript{5} & 84.00\textsubscript{3} & 77.27\textsubscript{3} \\
SU-NLP                 &  81.75\textsubscript{8} &  N/A &  N/A &  N/A \\
NoConflict           &  51.94\textsubscript{16} &  N/A &  N/A &  N/A \\
jitin                  &  67.39\textsubscript{14} & 70.49\textsubscript{8} & 52.23\textsubscript{13} & 62.05\textsubscript{12} \\
\hline
\textbf{ARC-NLP}                  &  81.35\textsubscript{9} & N/A & 81.73\textsubscript{5} & 72.42\textsubscript{8} \\
\textbf{ClassBases}                  &  82.30\textsubscript{6} & 80.78\textsubscript{1} & 85.39\textsubscript{2} & 73.48\textsubscript{5} \\
\textbf{colabhero}                 &  82.34\textsubscript{5} & 74.21\textsubscript{7} & 81.73\textsubscript{5} & 73.27\textsubscript{6} \\
\textbf{fine\_sunny\_day}                  &  85.00\textsubscript{2} & N/A & 80.74\textsubscript{8} & 82.45\textsubscript{1} \\
\textbf{gauravsingh}                  &  82.28\textsubscript{7} & 78.60\textsubscript{4} & 79.41\textsubscript{9} & 73.86\textsubscript{4} \\
\textbf{lizhuoqun2021\_iscas}                  &  85.12\textsubscript{1} & 80.01\textsubscript{2} & 85.87\textsubscript{1} & 81.19\textsubscript{2} \\
\hline
\end{tabular}}
\caption{The performance of the submissions in terms of F1-macro and their ranks as a subscript for each language and each team participating in CASE 2021 subtask 1. Bold teams indicate CASE 2022 entries.}
\label{CASE2021subtask1results}
\end{table}

\subsection{Subtask 2}
The extended results for CASE 2021 subtask 2 are provided in Table ~\ref{subtask2results}. The boldness indicates CASE 2022 entries.
ARC-NLP train an ensemble of transformers models using all training data to achieve 4th, 5th and 7th places in Spanish, English and Portuguese respectively. ClassBases finetune mLUKE-base for Portuguese and Spanish placing 5th in both, XLM-Roberta-large for English taking 8th place.\footnote{CamPros do not describe their model for subtask 2.}

\begin{table}[!th]
\centering
\scalefont{0.85}
\resizebox{\columnwidth}{!}{\begin{tabular}{lccc}
\hline
Team &   English & Portuguese & Spanish \\
\hline
ALEM                   &  79.67\textsubscript{9}  & 42.79\textsubscript{15} & 45.30\textsubscript{15} \\
AMU-EuraNova           &  75.64\textsubscript{14}  & 81.61\textsubscript{11} & 76.39\textsubscript{11} \\
DaDeFrTi               &  79.28\textsubscript{10}  & 86.62\textsubscript{6} & 85.17\textsubscript{6} \\
FKIE\_itf\_2021        &  64.96\textsubscript{16} & 75.81\textsubscript{13} & 70.49\textsubscript{14} \\
HSAIR &  78.50\textsubscript{11}  & 85.06\textsubscript{8} & 83.25\textsubscript{8} \\
IBM MNLP IE            &  84.56\textsubscript{4}  & 88.47\textsubscript{3} & 88.61\textsubscript{2} \\
IIITT                  &  82.91\textsubscript{7}  & 79.51\textsubscript{12} & 75.78\textsubscript{12} \\
SU-NLP                 &  83.05\textsubscript{6}  &  N/A &  N/A \\
NoConflict             &  85.32\textsubscript{3}  & 87.00\textsubscript{4} & 79.97\textsubscript{10} \\
jiawei1998             &  76.14\textsubscript{13}  & 84.67\textsubscript{9} & 83.05\textsubscript{9} \\
jitin                  &  66.96\textsubscript{15} & 69.02\textsubscript{14} & 72.94\textsubscript{13} \\
\hline
\textbf{ARC-NLP}                  &  83.77\textsubscript{5} & 86.53\textsubscript{7} & 87.20\textsubscript{4} \\
\textbf{CamPros}                 &  77.94\textsubscript{12} & 81.63\textsubscript{10} & 83.69\textsubscript{7} \\
\textbf{ClassBases}                  &  81.12\textsubscript{8} & 86.83\textsubscript{5} & 87.10\textsubscript{5} \\
\textbf{fine\_sunny\_day}                  &  85.75\textsubscript{2} & 89.67\textsubscript{1} & 88.78\textsubscript{1} \\
\textbf{lizhuoqun2021\_iscas}                  &  85.93\textsubscript{1} & 88.86\textsubscript{2} & 88.61\textsubscript{2} \\
\hline
\end{tabular}}
\caption{The performance of the submissions in terms of F1-macro and their ranks as a subscript for each language and each team participating in subtask 2. Bold teams indicate CASE 2022 entries.}
\label{subtask2results}
\end{table}

\subsection{Subtask 3}
The extended results for CASE 2021 subtask 3 are provided in Table ~\ref{subtask3results}. The boldness indicates CASE 2022 entries.
ARC-NLP achieve 1st place in both English and Spanish, 2nd place in Portuguese. They use an ensemble of English transformers models for English, Portuguese and Spanish test data. They train with only English data and translating Portuguese test data into English during model prediction. For Spanish test data, they train with English, translated Portuguese and translated Spanish, and test on translated Spanish data.

\begin{table}[!th]
\centering
\scalefont{0.85}
\resizebox{\columnwidth}{!}{\begin{tabular}{llll}
\hline
Team &   English & Portuguese & Spanish \\
\hline
DAAI                   &  80.40\textsubscript{4} & 90.23\textsubscript{6} & 81.83\textsubscript{6} \\
FKIE\_itf\_2021        &  77.05\textsubscript{7} & 91.33\textsubscript{4} & 82.52\textsubscript{4} \\
Handshakes AI Research &  79.01\textsubscript{5} & 90.61\textsubscript{5} & 81.95\textsubscript{5} \\
IBM MNLP IE            &  84.44\textsubscript{2} & 92.84\textsubscript{3} & 84.23\textsubscript{2} \\
NUS-IDS                &  81.20\textsubscript{3} & 93.03\textsubscript{1} & 83.15\textsubscript{3} \\
SU-NLP                 &  78.67\textsubscript{6} &  N/A &  N/A \\
\hline
\textbf{ARC-NLP}                &  85.11\textsubscript{1} & 93.00\textsubscript{2} & 85.25\textsubscript{1} \\
\hline
\end{tabular}}
\caption{The performance of the submissions in terms of CoNLL-2012 average score~\citet{Pradhan+14} and their ranks as a subscript for each language and each team participating in subtask 3. Bold teams indicate CASE 2022 entries.}
\label{subtask3results}
\end{table}

\subsection{Subtask 4}
The extended results for CASE 2021 subtask 4 are provided in Table ~\ref{subtask4results}. The boldness indicates CASE 2022 entries.
SPARTA employ two methods. Both of these methods build on pretrained XLM-Roberta-large and use a data augmentation technique (sentence reordering). For English and Portuguese, they gather articles that contain protest events from outside sources and use them for further pretraining. For Spanish, they use an XLM-Roberta-large model that was further pretrained on CoNLL 2002 Spanish data. They take 1st place both in Portuguese and Spanish, 3rd place in English.

ARC-NLP finetune an ensemble of transformers models for each language. They use all training data for Portuguese and Spanish, and only English for English test data. They achieve 2nd place in all languages.
EventGraph aim to solve event extraction as semantic graph parsing. They use a graph encoding method where the labels for triggers and arguments are represented as node labels, also splitting multiple triggers. They use the pretrained XLM-Roberta-large as their encoder. They achieve 4th place both in English and Portuguese, 5th place in Spanish.
ClassBases take 9th place in all languages finetuning XLM-Roberta-base transformers model.

\begin{table}[!th]
\centering
\scalefont{0.85}
\resizebox{\columnwidth}{!}{\begin{tabular}{llll}
\hline
{} & \multicolumn{3}{c}{Scores} \\
Team &   English & Portuguese & Spanish \\
\hline
AMU-EuraNova           & 69.96\textsubscript{7} & 61.87\textsubscript{8} & 56.64\textsubscript{8} \\
Handshakes AI Research & 73.53\textsubscript{5} & 68.15\textsubscript{6} & 62.21\textsubscript{6} \\
IBM MNLP IE            & 78.11\textsubscript{1} & 73.24\textsubscript{3} & 66.20\textsubscript{3} \\
SU-NLP                 &  2.58\textsubscript{10} &   N/A &   N/A \\
jitin                  & 66.43\textsubscript{8} & 64.19\textsubscript{7} & 58.35\textsubscript{7} \\
\hline
\textbf{ARC-NLP}                  & 77.83\textsubscript{2} & 73.84\textsubscript{2} &  	67.99\textsubscript{2} \\
\textbf{ClassBases}                  & 46.88\textsubscript{9} & 12.52\textsubscript{9} &  	37.09\textsubscript{9} \\
\textbf{EventGraph}                  & 74.76\textsubscript{4} & 71.72\textsubscript{4} & 64.48\textsubscript{5} \\
\textbf{SPARTA}                  & 76.60\textsubscript{3} & 74.56\textsubscript{1} & 69.86\textsubscript{1} \\
\textbf{lapardnemihk9989}                  & 72.18\textsubscript{6} & 70.98\textsubscript{5} & 64.83\textsubscript{4} \\
\hline
\end{tabular}}
\caption{The performance of the submissions in terms of F1 score based on CoNLL-2003~\citep{Sang+03} and their ranks as a subscript for each language and each team participating in subtask 4. Bold teams indicate CASE 2022 entries.}
\label{subtask4results}
\end{table}








\section{Conclusion}
\label{sec:conclusion}

The CASE 2022 extension consists of expanding the test data with more data in previously available languages, namely, English, Hindi, Portuguese, and Spanish, and adding new test data in Mandarin, Turkish, and Urdu for Sub-task 1, document classification. The training data from CASE 2021 in English, Portuguese and Spanish were utilized. Therefore, predicting document labels in Hindi, Mandarin, Turkish, and Urdu occurs in a zero-shot setting. 

The CASE 2022 workshop accepts reports on systems developed for predicting test data of CASE 2021 as well. We observe that the best systems submitted by CASE 2022 participants achieve between 79.71 and 84.06 F1-macro for new languages in a zero-shot setting. The winning approaches are mainly ensembling models and merging data in multiple languages. The best two submissions on CASE 2021 data outperform submissions from last year for Subtask 1 and Subtask 2 in all languages. Only the following scenarios were not outperformed by new submissions on CASE 2021: Subtask 3 Portuguese \& Subtask 4 English.

We aim at increasing number of languages and subtasks such as event coreference resolution~\cite{hurriyetoglu-et-al-2022-event} and event type classification\cite{hurriyetoglu-etal-2021-cross} in the scope of following edition of this shared task.

\bibliography{anthology,custom}

\begin{thebibliography}{21}
\expandafter\ifx\csname natexlab\endcsname\relax\def\natexlab#1{#1}\fi

\bibitem[{Duruşan et~al.(2022)Duruşan, Hürriyetoğlu, Yörük, Mutlu,
  Yoltar, Gürel, and Comin}]{durusan-et-al-2022-global}
Fırat Duruşan, Ali Hürriyetoğlu, Erdem Yörük, Osman Mutlu, Çağrı
  Yoltar, Burak Gürel, and Alvaro Comin. 2022.
\newblock \href {https://doi.org/10.48550/ARXIV.2206.10299} {Global contentious
  politics database (glocon) annotation manuals}.

\bibitem[{Fernandes et~al.(2022)Fernandes, Junior, da~Mata~Marques,
  da~Silva~Soares, and Filho}]{fernandes-etal-2022-ceia}
Diogo Fernandes, Adalberto Junior, Gabriel da~Mata~Marques, Anderson
  da~Silva~Soares, and Arlindo Rodrigues~Galvao Filho. 2022.
\newblock {CEIA-NLP at CASE 2022 Task 1: Protest News Detection for
  Portuguese}.
\newblock In \emph{Proceedings of the 5th Workshop on Challenges and
  Applications of Automated Extraction of Socio-political Events from Text
  (CASE 2022)}, online. Association for Computational Linguistics (ACL).

\bibitem[{H{\"u}rriyeto{\u{g}}lu et~al.(2021)H{\"u}rriyeto{\u{g}}lu, Mutlu,
  Y{\"o}r{\"u}k, Liza, Kumar, and Ratan}]{hurriyetoglu-etal-2021-multilingual}
Ali H{\"u}rriyeto{\u{g}}lu, Osman Mutlu, Erdem Y{\"o}r{\"u}k, Farhana~Ferdousi
  Liza, Ritesh Kumar, and Shyam Ratan. 2021.
\newblock \href {https://doi.org/10.18653/v1/2021.case-1.11} {Multilingual
  protest news detection - shared task 1, {CASE} 2021}.
\newblock In \emph{Proceedings of the 4th Workshop on Challenges and
  Applications of Automated Extraction of Socio-political Events from Text
  (CASE 2021)}, pages 79--91, Online. Association for Computational
  Linguistics.

\bibitem[{H{\"u}rriyeto{\u{g}}lu et~al.(2022)H{\"u}rriyeto{\u{g}}lu, Tanev,
  Zavarella, Yeniterzi, Mutlu, and
  Y{\"o}r{\"u}k}]{hurriyetoglu-etal-2022-challenges}
Ali H{\"u}rriyeto{\u{g}}lu, Hristo Tanev, Vanni Zavarella, Reyyan Yeniterzi,
  Osman Mutlu, and Erdem Y{\"o}r{\"u}k. 2022.
\newblock Challenges and applications of automated extraction of
  socio-political events from text (case 2022): Workshop and shared task
  report.
\newblock In \emph{Proceedings of the 5th Workshop on Challenges and
  Applications of Automated Extraction of Socio-political Events from Text
  (CASE 2022)}, online. Association for Computational Linguistics (ACL).

\bibitem[{Hürriyetoğlu et~al.(2022)Hürriyetoğlu, Mutlu, Beyhan, Duruşan,
  Safaya, Yeniterzi, and Yörük}]{hurriyetoglu-et-al-2022-event}
Ali Hürriyetoğlu, Osman Mutlu, Fatih Beyhan, Fırat Duruşan, Ali Safaya,
  Reyyan Yeniterzi, and Erdem Yörük. 2022.
\newblock \href {https://doi.org/10.48550/ARXIV.2203.10123} {Event coreference
  resolution for contentious politics events}.

\bibitem[{Hürriyetoğlu et~al.(2021)Hürriyetoğlu, Yörük, Mutlu, Duruşan,
  Yoltar, Yüret, and Gürel}]{hurriyetoglu-etal-2021-cross}
Ali Hürriyetoğlu, Erdem Yörük, Osman Mutlu, Fırat Duruşan, Çağrı
  Yoltar, Deniz Yüret, and Burak Gürel. 2021.
\newblock \href {https://doi.org/10.1162/dint_a_00092} {{Cross-Context News
  Corpus for Protest Event-Related Knowledge Base Construction}}.
\newblock \emph{Data Intelligence}, 3(2):308--335.

\bibitem[{Hürriyetoğlu et~al.(2020)Hürriyetoğlu, Zavarella, Tanev,
  Y{\"o}r{\"u}k, Safaya, and Mutlu}]{hurriyetoglu-etal-2020-automated}
Ali Hürriyetoğlu, Vanni Zavarella, Hristo Tanev, Erdem Y{\"o}r{\"u}k, Ali
  Safaya, and Osman Mutlu. 2020.
\newblock \href {https://www.aclweb.org/anthology/2020.aespen-1.1} {Automated
  extraction of socio-political events from news ({AESPEN}): Workshop and
  shared task report}.
\newblock In \emph{Proceedings of the Workshop on Automated Extraction of
  Socio-political Events from News 2020}, pages 1--6, Marseille, France.
  European Language Resources Association (ELRA).

\bibitem[{Kumari et~al.(2022)Kumari, Anand, Mohan, and
  Kumaraguru}]{kumari-etal-2022-campros}
Neha Kumari, Mrinal Anand, Tushar Mohan, and Ponnurangam Kumaraguru. 2022.
\newblock {CamPros at CASE 2022 Task 1: Transformer-based Multilingual Protest
  News Detection}.
\newblock In \emph{Proceedings of the 5th Workshop on Challenges and
  Applications of Automated Extraction of Socio-political Events from Text
  (CASE 2022)}, online. Association for Computational Linguistics (ACL).

\bibitem[{Mutlu(2022)}]{mutlu-2022-utilizing}
Osman Mutlu. 2022.
\newblock \href {https://doi.org/10.48550/ARXIV.2205.05468} {Utilizing
  coarse-grained data in low-data settings for event extraction}.

\bibitem[{Müller and Dafnos(2022)}]{muller-and-dafnos-2022-sparta}
Arthur Müller and Andreas Dafnos. 2022.
\newblock {SPARTA at CASE 2021 Task 1: Evaluating Different Techniques to
  Improve Event Extraction}.
\newblock In \emph{Proceedings of the 5th Workshop on Challenges and
  Applications of Automated Extraction of Socio-political Events from Text
  (CASE 2022)}, online. Association for Computational Linguistics (ACL).

\bibitem[{Palen-Michel et~al.(2022)Palen-Michel, Kim, and
  Lignos}]{palen-michel-etal-2022-multilingual}
Chester Palen-Michel, June Kim, and Constantine Lignos. 2022.
\newblock \href {https://aclanthology.org/2022.lrec-1.224} {Multilingual open
  text release 1: Public domain news in 44 languages}.
\newblock In \emph{Proceedings of the Thirteenth Language Resources and
  Evaluation Conference}, pages 2080--2089, Marseille, France. European
  Language Resources Association.

\bibitem[{Pradhan et~al.(2014)Pradhan, Luo, Recasens, Hovy, Ng, and
  Strube}]{Pradhan+14}
Sameer Pradhan, Xiaoqiang Luo, Marta Recasens, Eduard Hovy, Vincent Ng, and
  Michael Strube. 2014.
\newblock \href {http://www.aclweb.org/anthology/P14-2006} {Scoring coreference
  partitions of predicted mentions: A reference implementation}.
\newblock In \emph{Proceedings of the 52nd Annual Meeting of the Association
  for Computational Linguistics (Volume 2: Short Papers)}, pages 30--35,
  Baltimore, Maryland. Association for Computational Linguistics.

\bibitem[{Sahin et~al.(2022)Sahin, Ozcelik, Kucukkaya, and
  Toraman}]{sahin-etal-2022-arc-nlp}
Umitcan Sahin, Oguzhan Ozcelik, Izzet~Emre Kucukkaya, and Cagri Toraman. 2022.
\newblock {ARC-NLP at CASE 2022 Task 1: Ensemble Learning for Multilingual
  Protest Event Detection}.
\newblock In \emph{Proceedings of the 5th Workshop on Challenges and
  Applications of Automated Extraction of Socio-political Events from Text
  (CASE 2022)}, online. Association for Computational Linguistics (ACL).

\bibitem[{Suri et~al.(2022)Suri, Chopra, and Arora}]{suri-etal-2022-nsutnlp}
Manan Suri, Krish Chopra, and Adwita Arora. 2022.
\newblock {NSUT-NLP at CASE 2022 Task 1: Multilingual Protest Event Detection
  using Transformer-based Models}.
\newblock In \emph{Proceedings of the 5th Workshop on Challenges and
  Applications of Automated Extraction of Socio-political Events from Text
  (CASE 2022)}, online. Association for Computational Linguistics (ACL).

\bibitem[{Tan et~al.(2022{\natexlab{a}})Tan, Hürriyetoğlu, Caselli, Oostdijk,
  Hettiarachchi, Nomoto, Uca, and Liza}]{tan-etal-2022-event}
Fiona~Anting Tan, Ali Hürriyetoğlu, Tommaso Caselli, Nelleke Oostdijk, Hansi
  Hettiarachchi, Tadashi Nomoto, Onur Uca, and Farhana~Ferdousi Liza.
  2022{\natexlab{a}}.
\newblock Event causality identification with causal news corpus - shared task
  3, {CASE} 2022.
\newblock In \emph{Proceedings of the 5th Workshop on Challenges and
  Applications of Automated Extraction of Socio-political Events from Text
  (CASE 2022)}, Online. Association for Computational Linguistics.

\bibitem[{Tan et~al.(2022{\natexlab{b}})Tan, Hürrriyetoğlu, Caselli,
  Oostdijk, Nomoto, Hettiarachchi, Ameer, Uca, Liza, and
  Hu}]{tan-EtAl:2022:LREC}
Fiona~Anting Tan, Ali Hürrriyetoğlu, Tommaso Caselli, Nelleke Oostdijk,
  Tadashi Nomoto, Hansi Hettiarachchi, Iqra Ameer, Onur Uca, Farhana~Ferdousi
  Liza, and Tiancheng Hu. 2022{\natexlab{b}}.
\newblock \href {https://aclanthology.org/2022.lrec-1.246} {The causal news
  corpus: Annotating causal relations in event sentences from news}.
\newblock In \emph{Proceedings of the Language Resources and Evaluation
  Conference}, pages 2298--2310, Marseille, France. European Language Resources
  Association.

\bibitem[{Tjong Kim~Sang and De~Meulder(2003)}]{Sang+03}
Erik~F. Tjong Kim~Sang and Fien De~Meulder. 2003.
\newblock \href {https://doi.org/10.3115/1119176.1119195} {Introduction to the
  conll-2003 shared task: Language-independent named entity recognition}.
\newblock In \emph{Proceedings of the Seventh Conference on Natural Language
  Learning at HLT-NAACL 2003 - Volume 4}, CONLL ’03, page 142–147, USA.
  Association for Computational Linguistics.

\bibitem[{Wiriyathammabhum(2022)}]{wiriyathammabhum-2022-classbases}
Peratham Wiriyathammabhum. 2022.
\newblock {ClassBases at the CASE-2022 Multilingual Protest Event Detection
  Task: Multilingual Protest News Detection and Automatically Replicating
  Manually Created Event Datasets}.
\newblock In \emph{Proceedings of the 5th Workshop on Challenges and
  Applications of Automated Extraction of Socio-political Events from Text
  (CASE 2022)}, online. Association for Computational Linguistics (ACL).

\bibitem[{You et~al.(2022)You, Samuel, Touileb, and
  Øvrelid}]{you-etal-2022-eventgraph}
Huiling You, David Samuel, Samia Touileb, and Lilja Øvrelid. 2022.
\newblock {EventGraph at CASE 2021 Task 1: A General Graph-based Approach to
  Protest Event Extraction}.
\newblock In \emph{Proceedings of the 5th Workshop on Challenges and
  Applications of Automated Extraction of Socio-political Events from Text
  (CASE 2022)}, online. Association for Computational Linguistics (ACL).

\bibitem[{Yörük et~al.(2021)Yörük, Hürriyetoğlu, Duruşan, and Çağrı
  Yoltar}]{yoruk-etal-2021-random}
Erdem Yörük, Ali Hürriyetoğlu, Fırat Duruşan, and Çağrı Yoltar. 2021.
\newblock \href {https://doi.org/10.1177/00027642211021630} {Random sampling in
  corpus design: Cross-context generalizability in automated multicountry
  protest event collection}.
\newblock \emph{American Behavioral Scientist}, 0(0):00027642211021630.

\bibitem[{Zavarella et~al.(2022)Zavarella, Tanev, H{\"u}rriyeto{\u{g}}lu,
  Wiriyathammabhum, and De~Longueville}]{zavarella-etal-2022-covid19}
Vanni Zavarella, Hristo Tanev, Ali H{\"u}rriyeto{\u{g}}lu, Peratham
  Wiriyathammabhum, and Bertrand De~Longueville. 2022.
\newblock {Tracking COVID-19 protest events in the United States. Shared Task
  2: Event Database Replication, CASE 2022}.
\newblock In \emph{Proceedings of the 5th Workshop on Challenges and
  Applications of Automated Extraction of Socio-political Events from Text
  (CASE 2022)}, online. Association for Computational Linguistics (ACL).

\end{thebibliography}
\bibliographystyle{acl_natbib}




\end{document}